\documentclass[letterpaper, 10 pt, conference]{ieeeconf}  % Comment this line out if you need a4paper

\IEEEoverridecommandlockouts                              % This command is only needed if 
\usepackage{multirow}
\usepackage{graphicx}
\usepackage{array}
\usepackage{booktabs}
\usepackage{stmaryrd}
\usepackage{color}
\usepackage{arydshln}
\usepackage{makecell}
\usepackage{pifont}
\usepackage{tabularx}

\usepackage[table]{xcolor} 
\usepackage{tikz}
\definecolor{lightblue}{rgb}{0.8, 0.9, 1.0}

\usepackage{amsmath, amsfonts, amsthm, stackrel, amssymb}
\usepackage{float}
\usepackage{stfloats}
\usepackage{colortbl}
\usepackage{hyperref} 
\usepackage{bbding}  
\usepackage{array}    
\usepackage[utf8]{inputenc}
\usepackage{caption} 
\usepackage{cite}
\let\labelindent\relax
\usepackage{enumitem}
\usepackage{xspace}

\newcommand{\projname}{SurveilNav}

\title{\LARGE \bf
\projname: Collaborative Object Goal Navigation with Robot and Surveillance System}
\author{Anonymous Author}

\author{
Ming-Ming Yu$^{1,2}$, Qunbo Wang$^{3}$\textsuperscript{†}, Rongtao Xu$^{4}$, Yanghong Mei$^{2,5}$, Yirong Yang$^{1,2}$, \\ 
Longteng Guo$^{2,5}$, Wenjun Wu$^{1,6}$, and Jing Liu$^{2,5}$\textsuperscript{†}
\thanks{$^{1}$Beihang University; $^{2}$Institute of Automation, Chinese Academy of Sciences; $^3$Beijing Jiaotong University; $^4$ATeam;
$^{5}$University of Chinese Academy of Sciences; $^6$Hangzhou International Innovation Institute, Beihang University.   \quad$^{\dagger}$Corresponding authors. Emails: {\tt\small mingmingyu@buaa.edu.cn}, {\tt\small wangqb6@outlook.com}}
}

\begin{document}

\maketitle
\thispagestyle{empty}
\pagestyle{empty}

%%%%%%%%%%%%%%%%%%%%%%%%%%%%%%%%%%%%%%%%%%%%%%%%%%%%%%%%%%%%%%%%%%%%%%%%%%%%%%%%
\begin{abstract}
With the growing deployment of surveillance systems in factories, offices, and homes, integrating them with robots offers a promising direction for collaborative and efficient task execution. However, existing approaches largely focus on single-robot scenarios and struggle with multi-view collaboration in large-scale environments.
In this paper, we present a novel indoor collaborative object navigation dataset built on Habitat-Sim, featuring 206 cameras across 74 floors. The dataset enables systematic evaluation of an agent’s ability to exploit multi-view surveillance information. 
To address the limitations of single-robot perception, we propose \projname, a collaborative navigation framework that integrates active camera scheduling, joint 2D/3D mapping, VLM-based value estimation, and collaborative target verification. 
By synergizing the robot's dynamic local perception with the static global view of surveillance, this architecture effectively overcomes both the limited perception range of single agents and the inherent blind spots of fixed cameras, resolving inefficient exploration.
Experimental results on the HM3D dataset demonstrate that \projname \   substantially outperforms existing methods, achieving state-of-the-art performance in both exploration efficiency and navigation success rate. Moreover, the system shows strong potential for applications in large-scale search, home environments, and rescue missions.
\end{abstract}

\section{Introduction}
 Object goal navigation (ObjectNav) requires an agent to automatically locate a specified object in previously unseen environments, posing a fundamental challenge for embodied intelligence. With the rapid advancement of large language models (LLMs) and vision-language models (VLMs), zero-shot ObjectNav has made significant progress.
These approaches typically construct explicit environmental maps and leverage LLMs or VLMs to reason over them, selecting valuable frontiers or candidate waypoints for exploration. Recent works~\cite{gadre2023cows, zhou2023esc, yu2023l3mvn, yokoyama2024vlfm, wu2024voronav} have demonstrated remarkable improvements in exploration efficiency. 
By leveraging the extensive prior knowledge within LLMs, these methods achieve performance comparable to models trained on massive navigation trajectories.
Compared with data-driven approaches, they often exhibit superior generalization to novel environments. 

Nonetheless, a key limitation remains: nearly all existing methods are confined to single-robot systems, without leveraging multi-sensor or cross-view collaboration.
In real-world scenarios, a single egocentric viewpoint often provides limited coverage. This problem becomes more pronounced in large-scale, multi-floor environments, where both exploration efficiency and accuracy drop significantly. A similar issue has long been recognized in autonomous driving: single-vehicle perception is insufficient for safe decision-making. To address this, researchers have developed collaborative perception, where vehicles, roadside units, and infrastructure share information to expand perceptual coverage and enhance system accuracy and robustness~\cite{li2022v2x, xu2022opv2v, yu2022dair}. Inspired by this, we argue that indoor navigation can also benefit from robot–surveillance collaboration. While collaborative perception has been explored in autonomous driving, systematic exploration of how to effectively coordinate robots with surveillance systems in indoor settings remains lacking.

\begin{figure}[t]
  \centering
  \includegraphics[width=0.48\textwidth]{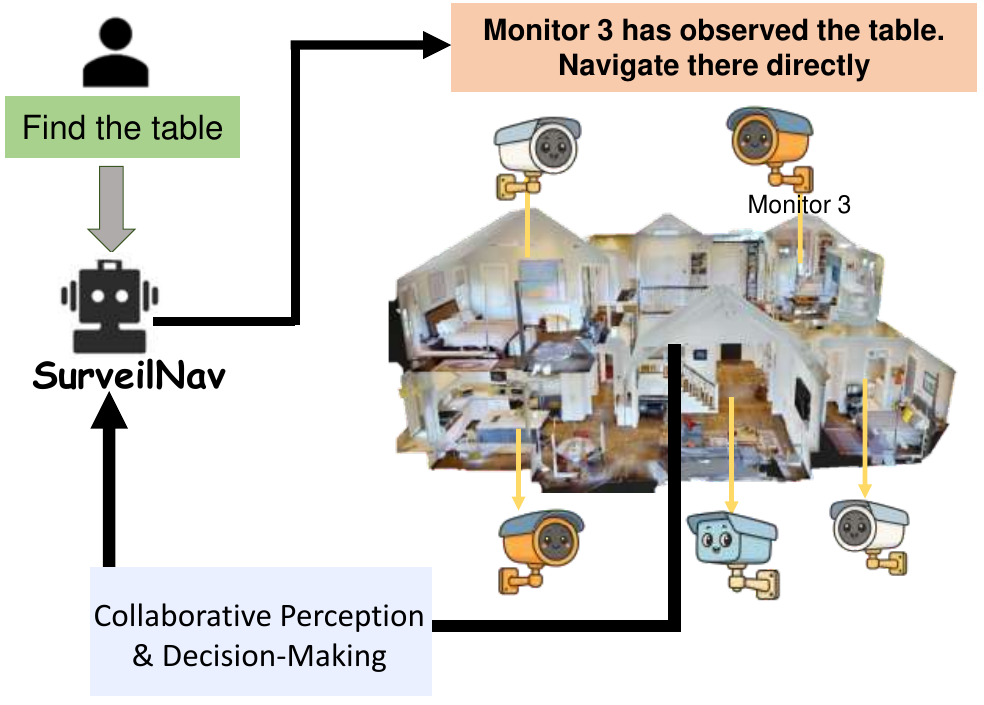}
  \caption{\projname\ workflow. Monitor~\#3 detects the target table and provides direct navigation guidance, enabling the agent to efficiently localize and approach the target through collaborative perception and decision-making.}
  \label{fig:collab_flow}
    \vspace{-0.64cm}
\end{figure}

To bridge this gap, we introduce a new research task: collaborative object-goal navigation with robots and surveillance systems. 
For this purpose, we construct the first dataset on Habitat-Sim, spanning 36 scenes, 74 floors, and 206 surveillance cameras. 
As illustrated in Figure~\ref{fig:collab_flow}, agents are required not only to explore using their onboard perception, but also to actively query surveillance viewpoints to enhance efficiency and accuracy in target localization. 
This formulation mirrors real-world scenarios where robots collaborate with infrastructure to accomplish complex tasks.

Building on this dataset, we introduce \textbf{\projname}, a novel collaborative object navigation framework. Our framework integrates active camera invocation, joint 3D map construction, VLM-based semantic value estimation, and target verification mechanisms. 
Specifically, the agent dynamically coordinates with the surveillance network to decide whether to directly approach a localized target or conduct autonomous exploration guided by collaborative perception. 
By leveraging this integrated architecture, \projname\ achieves a powerful synergy between global and local perception. It not only breaks through the bottleneck of limited perception range in single robots and compensates for the inherent blind spots of fixed surveillance, but also reduces false detections through multi-view validation. Ultimately, this comprehensive approach substantially improves both exploration efficiency and overall navigation performance.

Experimental results on the HM3D dataset demonstrate that \projname \   achieves state-of-the-art performance in large-scale and multi-floor scenarios, significantly surpassing existing methods in both exploration efficiency and navigation success rate. Moreover, the system shows strong potential in real-world applications such as large-scale search, household scenarios, and rescue missions where efficient collaborative perception is critical. 

Our main contributions are as follows: (1)We propose the first indoor object search dataset for robot–surveillance collaboration, covering 36 scenes, 74 floors, and 206 cameras, providing a new benchmark for multi-view navigation. (2)We introduce \projname, a collaborative object navigation framework based on VLMs, enabling agents to actively invoke surveillance systems for joint map construction, semantic exploration, and target verification. (3)We conduct large-scale experiments on HM3D, showing that \projname \ significantly outperforms existing methods in both exploration efficiency and navigation success rate, while demonstrating strong practicality.

\section{Related Work}

\subsection{Visual Navigation}
Visual navigation is a fundamental task in embodied intelligence.
% , integrating perception and planning to accomplish specific objectives. 
Recent advancements have introduced a variety of visual navigation tasks, such as point navigation~\cite{wijmans2019dd, savva2019habitat, chaplot2020object}, image-goal navigation~\cite{zhu2017target}, object-goal navigation~\cite{batra2020objectnav}, and vision-language navigation~\cite{krantz2020beyond, zhang2024navid, zheng2024towards, mei2025urbannav}. Point navigation tasks utilize coordinates relative to the robot's starting point as the target, while image-goal navigation aims for a target image. In vision-language navigation, agents follow step-by-step instructions to reach the target location. Compared to these tasks, object-goal navigation employs object category names as targets, demanding more robust exploration capabilities than vision-language navigation tasks. Thus, we focus on the object navigation task within unknown environments.

\subsection{Object Goal Navigation}

Existing approaches to object-goal navigation can be broadly divided into two categories. 
The first class of methods leverages pre-trained vision or language models as backbones, and trains navigation policies with extensive navigation trajectories via supervised or reinforcement learning~\cite{zhu2017target,ramrakhya2022habitat, zhang2024uni, zeng2024poliformer,yu2025c}. 
While effective, these approaches face two key limitations: they are confined to the finite set of object categories seen during training, limiting generalization to open environments, and the simulation-to-reality gap reduces the transferability of task-specific training to real-world scenarios.  
To overcome these challenges, recent work has shifted toward Zero-Shot Object Navigation~\cite{kuang2024openfmnav,gadre2023cows, yokoyama2024vlfm,wu2024voronav,sun2024prioritized,yu2025ranger}. 
These methods explicitly build environmental maps and employ vision-language models (VLMs)~\cite{achiam2023gpt, bai2023qwen} or large language models (LLMs)~\cite{grattafiori2024llama} for reasoning, selecting informative frontiers or waypoints for exploration. 
For instance, COW~\cite{gadre2023cows} guides the robot to explore the nearest frontier until the target is detected using CLIP features~\cite{radford2021learning} and open-vocabulary object detectors~\cite{caron2021emerging}. 
ESC~\cite{zhou2023esc}, L3MVN~\cite{yu2023l3mvn}, and VoroNav~\cite{wu2024voronav} enhance decision-making with LLMs, while VLFM~\cite{yokoyama2024vlfm} uses a VLM to assign semantic values to frontiers based on egocentric observations and textual prompts.  
To further enhance efficiency, SEEK~\cite{ginting2024seek} utilizes spatial priors like floor plans, while GOAT~\cite{chang2023goat} leverages accumulated memory from past tasks.
Despite these advances, most approaches remain limited to single-robot systems, though some efforts have begun to explore cross-agent collaboration, such as air-ground robotic teams~\cite{miller2022stronger}. In contrast, we propose a collaborative framework that couples mobile robots with a surveillance system to enhance task efficiency.

\begin{figure*}[!ht]
  \centering
  \includegraphics[width=1.0\textwidth]{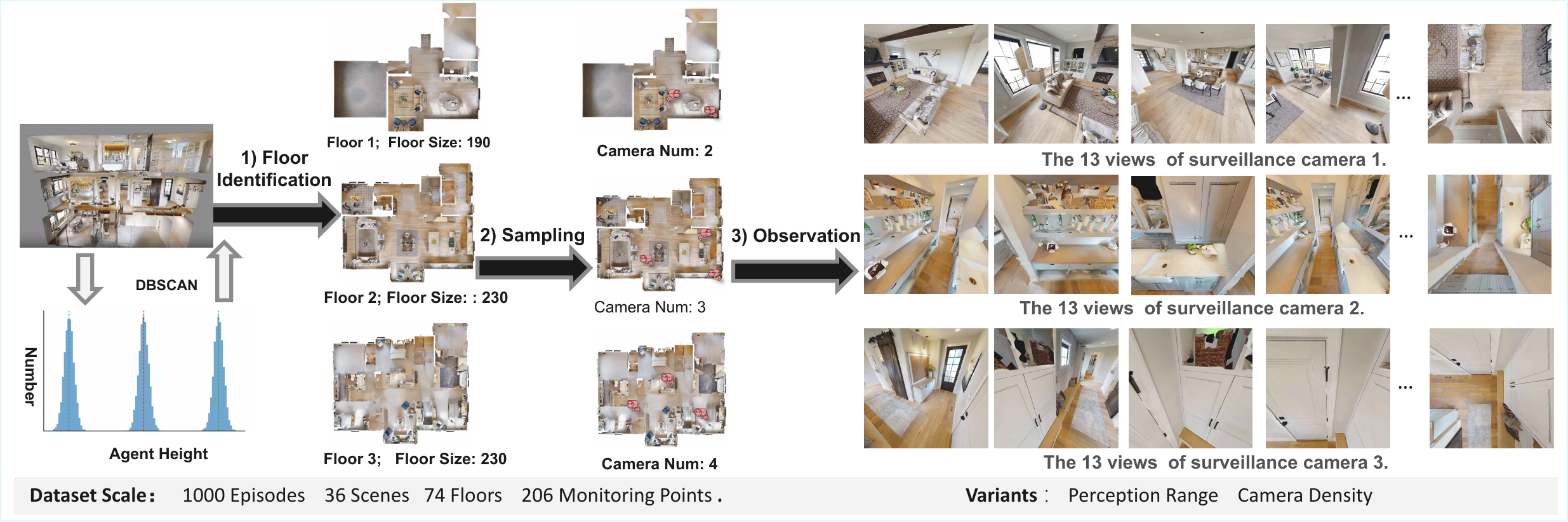}
\caption{
The surveillance camera observation generation pipeline, consisting of (a) floor identification, (b) camera sampling, and (c) observation configuration.
}
 \label{fig:data_gen}
 \vspace{-0.5cm}
\end{figure*}

\subsection{Collaborative perception in autonomous driving}

Autonomous driving has attracted significant attention in recent years, but standalone systems often suffer from limited perception ranges, which may lead to safety risks. To overcome this, \textit{collaborative perception} leverages information from multiple vehicles, infrastructure, and roadside units to expand the field of view and improve accuracy. With advances in deep learning and the release of large-scale datasets such as V2X-Sim~\cite{li2022v2x}, OPV2V~\cite{xu2022opv2v}, and DAIR-V2X~\cite{yu2022dair}, research in this field has accelerated. Collaborative perception methods can be categorized into early (e.g., Cooper~\cite{chen2019cooper}, Coop3D~\cite{arnold2020cooperative}), intermediate (e.g., V2VNet~\cite{wang2020v2vnet}), and late collaboration (e.g., OptiMatch~\cite{song2023cooperative}), each improving robustness through tailored communication, feature fusion, and optimization strategies. However, these efforts have focused almost exclusively on outdoor driving scenarios, leaving open the question of how to effectively coordinate robots and surveillance systems in indoor environments.

\section{TASKS AND DATASETS}
\subsection{Task Definition}
\label{sec:task_definition}

In a surveillance and agent collaborative semantic navigation task, the agent needs to rely on its first-person perspective observations while collaborating with third-person perspective observations from surveillance systems deployed throughout the building. The goal is to search for a specified object category in a previously unseen environment without prior mapping.
Formally, the scene set is denoted as $\mathcal{S} = \{s_1, \dots, s_k\}$, and the category set is denoted as $\mathcal{C} = \{c_1, \dots, c_m\}$. For a given scene $s_i \in \mathcal{S}$, the deployed surveillance system is defined as $\mathcal{H}_i = \{h_1, h_2, \dots, h_{n_i}\}$, where $n_i$ represents the specific number of cameras in that scene. 
For each episode, the agent is randomly initialized in an unknown scene $s_i$ and must locate a target category $c \in \mathcal{C}$. At each time step $t$, the robot acquires a first-person observation $O_t$, which includes an RGB-D image, the robot's pose (position and orientation), the target category $c$, as well as the observations and poses of all cameras in the surveillance system $\mathcal{H}_i$. 
The action space $\mathcal{A}$ consists of \textit{move forward} (0.25m), \textit{turn left} ($30^{\circ}$), \textit{turn right} ($30^{\circ}$), \textit{look up} ($30^{\circ}$), \textit{look down} ($30^{\circ}$), and \textit{stop}. An episode is deemed successful if the agent executes the \textit{stop} action when its geodesic distance to the target object is sufficiently small, with a maximum episode length of 500 steps.

\subsection{\projname \    Dataset Construction}

We develop the \projname \ dataset for collaborative semantic navigation based on Habitat HM3Dv2~\cite{yadav2023habitat}. As illustrated in Fig.~\ref{fig:data_gen}, the construction pipeline consists of three phases:
\textbf{(1) Floor Identification}: The agent is first placed within the scene, and feasible points are recorded. 
DBSCAN clustering is then applied to the agent’s height coordinates to segment and identify different floor levels.  
\textbf{(2) Camera Sampling}: For each floor, the number of surveillance cameras is determined according to the floor area, with one camera allocated per $100 \,\text{m}^2$ and placed at a height of 1.8 meters. 
During sampling, candidate positions with an excessive proportion of black pixels (caused by artifacts in the 3D scans) are skipped. 
To ensure optimal coverage, the final placements are refined using farthest point sampling.  
\textbf{(3) Observation Generation}: To simulate realistic, wide-area monitoring, each camera records 13 viewpoints: 12 surrounding views with a $-30^{\circ}$ tilt (at $30^{\circ}$ azimuth intervals) and one $-90^{\circ}$ nadir view. Each viewpoint captures $1280 \times 1280$ RGB-D images and precise poses. While real-world surveillance often lacks depth sensors, modern monocular estimation (e.g., DepthAnything~\cite{yang2024depth}) has demonstrated the feasibility of metric depth recovery. Thus, we utilize ground-truth depth from Habitat-Sim as an empirical upper-bound to isolate the core effectiveness of our collaborative framework from potential depth-estimation noise.
Following this pipeline, we generate 1,000 episodes across 36 scenes, 74 floors, and 206 camera placements. 
In line with the standard ObjNav task definition, the dataset includes six object goal categories: chair, couch, potted plant, bed, toilet, and TV.  
To further analyze the impact of infrastructure density and visibility, we provide additional variants with varying camera densities (one per $200 \,\text{m}^2$ vs. $100 \,\text{m}^2$) and field-of-view configurations (panoramic vs. half views).

\section{Methodology}

\subsection{Overview}

\begin{figure*}[htpb]
  \centering
  \includegraphics[width=0.95\textwidth]{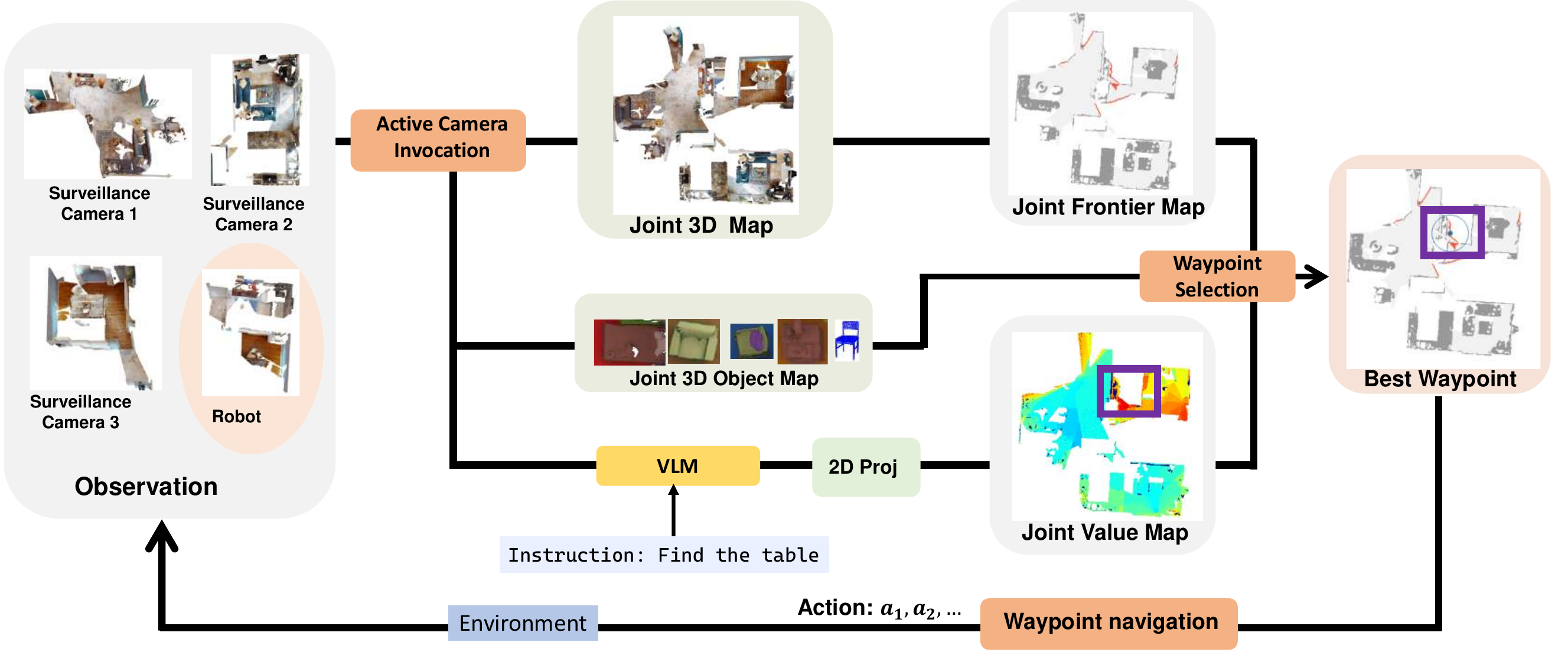}
  \caption{The proposed system, \projname, consists of several key components: active camera invocation, joint 3D map construction through multi-source observation alignment, joint value map generation with a vision-language model (VLM), joint 3D object confirmation, and waypoint selection and navigation.}
  \label{fig:framework}
  \vspace{-0.3cm}
\end{figure*}

As illustrated in Fig.~\ref{fig:framework}, at each time step $t$, \projname \ actively selects surveillance cameras based on its current position $\mathbf{p}_t$ and the distribution of monitoring locations. The system then constructs a unified 3D map representation by integrating egocentric observations from the robot with third-person visual perspectives from multiple cameras. Building on this representation, a joint 2D frontier map is generated to separate explored regions from unexplored areas. Meanwhile, the VLM assesses the relevance between multi-source observations and the task objective, and projects the resulting importance weights onto a 2D plane to form a joint value map.
In parallel, the system fuses multi-view detections and semantic information to construct a unified 3D object map, where a confidence-based fusion mechanism ensures accurate target perception. Finally, the system selects the optimal waypoint from candidate objects and frontiers according to the value map, enabling the robot to efficiently achieve object-goal navigation.

\subsection{Active Camera Invocation}

To optimize navigation efficiency and minimize computational overhead, the agent dynamically invokes a subset of surveillance cameras, $\mathbf{H}_{t} \subseteq \mathcal{H}_i$, based on spatial relevance. Let $\mathbf{p}_t = [x_t, y_t, z_t^a]^\top$ denote the agent's global pose at time step $t$, where $z_t^a$ represents the altitude of the agent's camera center. Similarly, let $\mathbf{p}_j = [x_j, y_j, z_j^c]^\top$ denote the position of surveillance camera $h_j \in \mathcal{H}_i$, where $z_j^c$ is its fixed installation height.
To ensure floor-level consistency, we estimate the absolute elevation of the supporting floor plane for both the agent and the cameras. Specifically, let $\hat{z}_t^a = z_t^a - \Delta z^a$ and $\hat{z}_j^c = z_j^c - \Delta z^c$ represent the inferred global altitude of the floor surface where the agent and camera $h_j$ are situated, respectively. Here, $\Delta z^a$ and $\Delta z^c$ denote the known vertical mounting offsets of the sensors relative to their respective local floors. The activation state of camera $h_j$, denoted by $\alpha_{j,t} \in \{0, 1\}$, is determined by a vertical alignment constraint:
\begin{equation}
\alpha_{j,t} = 
\begin{cases} 
1, & \text{if } |\hat{z}_t^a - \hat{z}_j^c| \leq \tau_z \\
0, & \text{otherwise}
\end{cases},
\end{equation}
where $\tau_z$ is a predefined threshold for vertical displacement. The active camera set at time $t$ is formally defined as $\mathbf{H}_{t} = \{h_j \in \mathcal{H}_i \mid \alpha_{j,t} = 1\}$. This mechanism effectively filters out irrelevant observations from different floors, thereby reducing the computational complexity for subsequent multi-source fusion and ensuring the scalability of the collaborative system in multi-floor environments.

\subsection{Map Representation} 

\subsubsection{Joint 3D Point Cloud Map Construction}

In the joint 3D point cloud map construction, the robot acquires local point clouds $\mathcal{P}_t$ through its RGB-D sensor, while the surveillance system provides point clouds $\mathcal{P}_j$ from fixed cameras. The robot's local point cloud is transformed into the global coordinate frame using its current pose $T_t = [R_t | \mathbf{t}_t]$, i.e., $\mathbf{p}_t^{\text{global}} = R_t \cdot \mathbf{p}_t + \mathbf{t}_t$, where $\mathbf{p}_t$ represents the point cloud coordinates in the local frame. 
Similarly, the surveillance point cloud $\mathcal{P}_j$ is transformed into the global frame using its fixed pose $T_j = [R_j | \mathbf{t}_j]$. By aligning and merging $\mathbf{p}_t^{\text{global}}$ and $\mathbf{p}_j^{\text{global}}$, a joint 3D point cloud map is constructed.

\subsubsection{2D Map Construction}
Once the unified 3D point cloud map is constructed, we project its points onto a 2D plane to derive two complementary representations: an obstacle map and an exploration map. The obstacle map is obtained by projecting points located above the floor level, whereas the exploration map is generated using all available 3D points.
To determine the frontier regions, we enlarge the obstacle boundaries through morphological dilation and then subtract the exploration map from the obstacle map. The resulting frontier map marks the interface between explored and unexplored areas. During navigation, both the distribution and the number of frontiers are continuously updated. When the robot has fully explored the environment, these frontiers naturally disappear.

\subsubsection{Joint Value Map}
The joint value map is designed to quantify the semantic relevance of each location in the explored area to the target object, and it has the same shape as the exploration map. For both surveillance images and the robot’s onboard observations, we employ CLIP to evaluate task-related relevance. Specifically, we compute the cosine similarity between the RGB image features $f_v$ and a text prompt feature $f_t$ formulated as ``seems like there is a [object name]'', i.e., $\cos(f_v, f_t)$.
By leveraging depth information together with the camera’s pose (position and orientation), these relevance scores are projected into the top-down map space. To integrate information from multiple viewpoints, we adopt an averaging fusion strategy. 
Formally, for each grid cell $(i,j)$, the joint relevance score is updated as:
\begin{equation}
s^{\text{joint}}(i,j) \leftarrow \frac{s^{\text{surveil}}(i,j) + s^{\text{robot}}(i,j)}{2},
\label{eq:joint_value}
\end{equation}
where $s^{\text{surveil}}(i,j)$ and $s^{\text{robot}}(i,j)$ denote the relevance values derived from the surveillance system and the robot’s local observations, respectively.

\subsection{Joint 3D Object Map and Target Confirmation}

\begin{figure*}[htpb]
  \centering
  \includegraphics[width=0.82\textwidth]{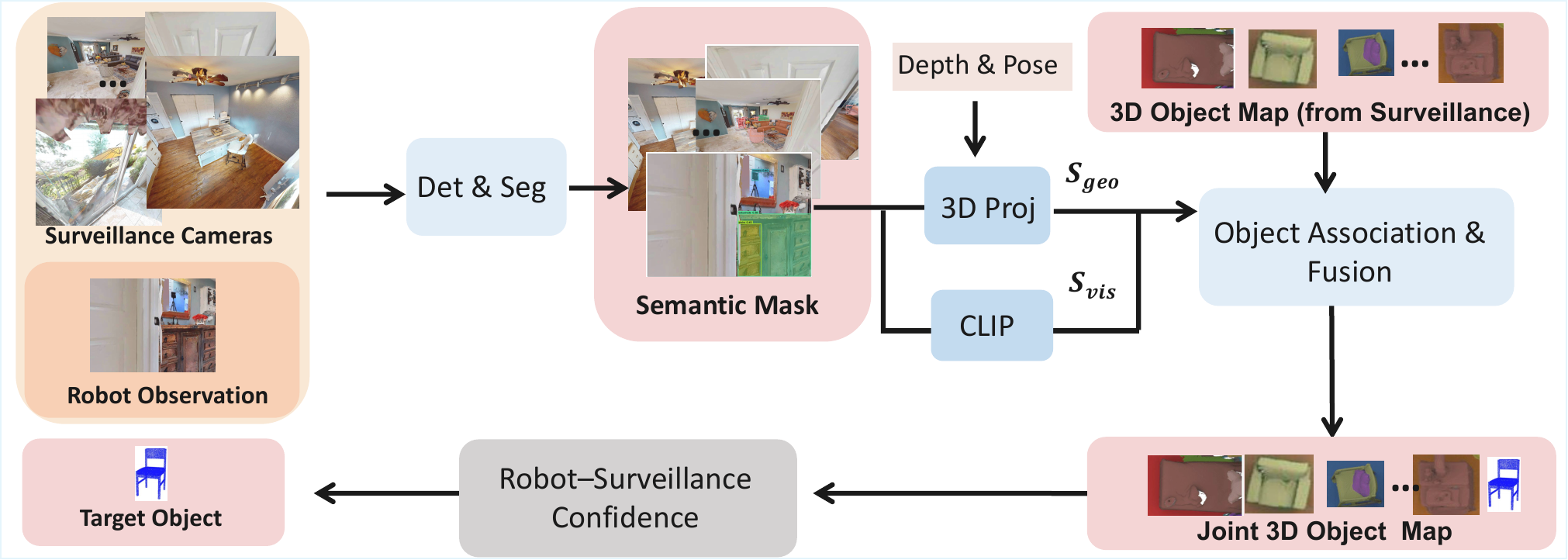}
    \caption{The process of constructing the joint 3D object map and confirming the target.}
  \label{fig:object_map}
  \vspace{-0.40cm}
\end{figure*}

\subsubsection{Joint 3D Object Map Construction}

As shown in Figure~\ref{fig:object_map}, we process observations captured by both the robot and the surveillance cameras. Specifically, Grounding DINO is adopted as an open-vocabulary detector, and MobileSAM~\cite{kirillov2023segment} is applied to produce object masks. Using depth information, these masks are projected back into 3D space to construct point cloud representations. For each mask, the corresponding image region is cropped according to its bounding box, and CLIP is employed to extract semantic features. This procedure yields a semantic point cloud enriched with confidence values, category labels, and feature embeddings.

\textbf{Joint Object association. } For each newly detected object $i$ from either the \textbf{robot} or a \textbf{surveillance camera}, we compute both geometric and semantic similarity with respect to all existing objects in the joint map. The \textit{geometric similarity} $S_{\text{geo}}(i, j) = \text{IoU}(\text{PointCloud}_i, \text{PointCloud}_j)$ is calculated as the Intersection over Union (IoU) of the point clouds between the newly detected object and an existing object. The \textit{semantic similarity} $S_{\text{vis}}(i, j) = \text{Sim}(\text{SemanticFeature}_i, \text{SemanticFeature}_j)$ is computed from the semantic feature embeddings of the newly detected object and the existing objects.
The overall similarity score between a new detection $i$ (from the robot or a surveillance camera) and an existing object $j$ is defined as a weighted sum of the two similarities:
\begin{equation}
s(i, j) = w_1 S_{\text{vis}}(i, j) + w_2 S_{\text{geo}}(i, j).
\end{equation}
If this score is lower than a threshold $\tau$, the system registers a new object instance; otherwise, the detection is matched with the existing object with the highest similarity.

\textbf{Joint Object fusion. } 
For objects successfully associated between the \textbf{robot} and \textbf{surveillance cameras}, we merge their point clouds into a unified representation. The corresponding features are then updated using a weighted average that reflects the detection frequencies from both sources.

\subsubsection{Joint Object Confirmation}

In this phase, candidate objects extracted from the semantic point cloud are evaluated based on their object confidence.
For each object, the confidence score is defined as the maximum value between the local observations from the robot and the global observations from the surveillance cameras. The category of the object is then determined by the class associated with this maximum confidence.
If the resulting object confidence surpasses a predefined threshold, the candidate is confirmed as the target.
As the robot moves and integrates additional observations, these confidence scores are dynamically updated, enabling more reliable confirmation.
If, during this process, the predicted category of an object changes, the detection is regarded as a false positive, which often arises from limited or suboptimal viewpoints of the surveillance system.
This confidence-evolution mechanism allows static wide-range observations from surveillance cameras and dynamic local observations from the robot to complement each other, leading to more robust and consistent object confirmation.

\subsection{Waypoint Selection and Navigation} 

After initialization, the robot adaptively selects waypoints for navigation according to the detection status of target objects.
If a target object has been confirmed, its location is directly assigned as the waypoint.
Otherwise, the robot chooses either a promising low-confidence candidate object or the highest-value frontier as the exploration waypoint.
To reach the designated waypoint from its current position, the robot employs the Fast Marching Method (FMM)~\cite{sethian1996fast} as the local planner.
The planner iteratively selects feasible local goals within the robot’s vicinity and generates corresponding actions to approach the target step by step.
At each iteration, both the map and the waypoint are updated with real-time observations, enabling adaptive and accurate navigation.

\begin{table*}[htbp]
    \centering
    \small
    \caption{Comparison with single-robot navigation methods on the HM3D dataset (results from MCoCoNav~\cite{shen2025enhancing}).}
    \begin{tabular}{p{4.2cm} c c p{4.5cm} c c}
        \toprule
        \multirow{2}{*}{Method} & \multirow{2}{*}{Zero-Shot} & \multirow{2}{*}{Training-Free} & \multirow{2}{*}{LLM/VLM} & \multicolumn{2}{c}{HM3D-v0.2} \\
        \cmidrule(lr){5-6}
        & & & & SPL$\uparrow$ & SR$\uparrow$ \\
        \midrule
        Random Walking & \checkmark & \checkmark & None & 0.000 & 0.000 \\
        Frontier Based~\cite{sethian1996fast} & \checkmark & \checkmark & None & 12.3 & 23.7 \\
        Random Samples & \checkmark & \checkmark & None & 14.3 & 30.0 \\
        \midrule
        VLFM~\cite{yokoyama2024vlfm} & \checkmark & \ding{55} & BLIP & 32.7 & 64.0 \\
        \midrule
        L3MVN~\cite{yu2023l3mvn} & \checkmark & \checkmark & RoBERTa-large  & 23.1 & 50.4 \\
        ESC~\cite{zhou2023esc} & \checkmark & \checkmark  & GPT-3.5  & 22.3 & 39.2 \\
        Single-NavGPT~\cite{yu2023co}  & \checkmark & \checkmark & GPT-3.5  & 21.5 & 53.9 \\
        VoroNav~\cite{wu2024voronav}  & \checkmark & \checkmark & GPT-3.5 & 26.0 & 42.0 \\
        OpenFMNav~\cite{kuang2024openfmnav}  & \checkmark & \checkmark & GPT-4/GPT-4V  & 24.4 & 54.9 \\
        % SGNav~\cite{}  & \checkmark & \checkmark & GPT-4  & 49.6 & 25.5
        MCoCoNav~\cite{shen2025enhancing}  & \checkmark & \checkmark & GLM-4V  &29.7 & 63.4 \\
        InstructNav~\cite{long2024instructnav} & \checkmark & \checkmark & Linguistic & 20.9 & 58.0\\
        VLN-Game~\cite{yu2024vln}  & \checkmark & \checkmark & CLIP & 26.9 & 66.7 \\
        \midrule
        \rowcolor{gray!20}
        Ours (FMM, w/o Surveillance) & \checkmark & \checkmark & CLIP & 26.1 & 62.7 \\
        \rowcolor{gray!20}
        Ours (FMM, with Surveillance) & \checkmark & \checkmark & CLIP & 34.5 & 67.4 \\
        \rowcolor{gray!20}
        Ours (SP, w/o Surveillance) & \checkmark & \checkmark & CLIP & 28.0 & 68.7 \\
        \rowcolor{gray!20}
        Ours (SP, with Surveillance) & \checkmark & \checkmark & CLIP & \textbf{36.4}  & \textbf{71.1}\\
        \bottomrule
    \end{tabular}
    \label{tab:hm3d_comparison}
    \vspace{-0.35cm}
\end{table*}

\section{Experiments}

\subsection{Experimental Setup.}

\textbf{Metrics.} We use navigation success rate (SR) and success rate weighted by navigation path length (SPL) as evaluation metrics. SR represents the percentage of successful episodes out of the total episodes. SPL measures the efficiency of reaching the goal in addition to the success rate.

\textbf{Implementation Details.} 
In the target association module, the similarity threshold for matching two targets is set to 1.35. The agent captures images at a resolution of 640×480 with a field of view (FOV) of 79 degrees, while the surveillance camera captures images at a resolution of 1280×1280 with an FOV of 100 degrees. For the active invocation module, the threshold for determining whether the surveillance camera and the robot are on the same floor is set to 0.4 meters. For the 2D map, the resolution is configured to 5 cm, with a map size of 2400 cm. 
For point cloud processing, downsampling uses a 2.5 cm voxel grid, and noise is removed using DBSCAN with an epsilon value of 0.05 and a minimum point count of 10. 
All point cloud operations are implemented using Open3D.

\subsection{Comparison with Single-Robot Navigation Methods}

\textbf{Comparison with Single-Robot Navigation Methods.} 
Table~\ref{tab:hm3d_comparison} summarizes the performance of our method on the HM3D dataset. 
The key advantage of \projname \   lies in its collaborative mechanism between surveillance cameras and the robot, which significantly improves navigation compared to single-robot systems. 
With the Shortest Path (SP) planner, our method achieves an SPL of 36.4 and an SR of 71.1, while the Fast Marching Method (FMM) attains 34.5 SPL and 67.4 SR. 
These results outperform the best baseline, MCoCoNav (29.7 SPL, 63.4 SR), by +6.7 in SPL and +7.7 in SR, and also exceed VLN-Game.

\textbf{Impact of Surveillance Inputs.} 
Removing surveillance inputs leads to a clear performance drop. 
For FMM, the SPL/SR decreases from 34.5/67.4 to 26.1/62.7, while SP drops from 36.4/71.1 to 28.0/68.7. 
These results highlight the effectiveness of surveillance in enhancing both navigation efficiency and success. 
Surveillance cameras provide a global and static view of the environment, while the robot contributes local sensing and active exploration. 
Together, they yield complementary strengths that enable more robust and efficient navigation.

\begin{figure*}[htbp]
  \centering
  \includegraphics[width=0.9\textwidth]{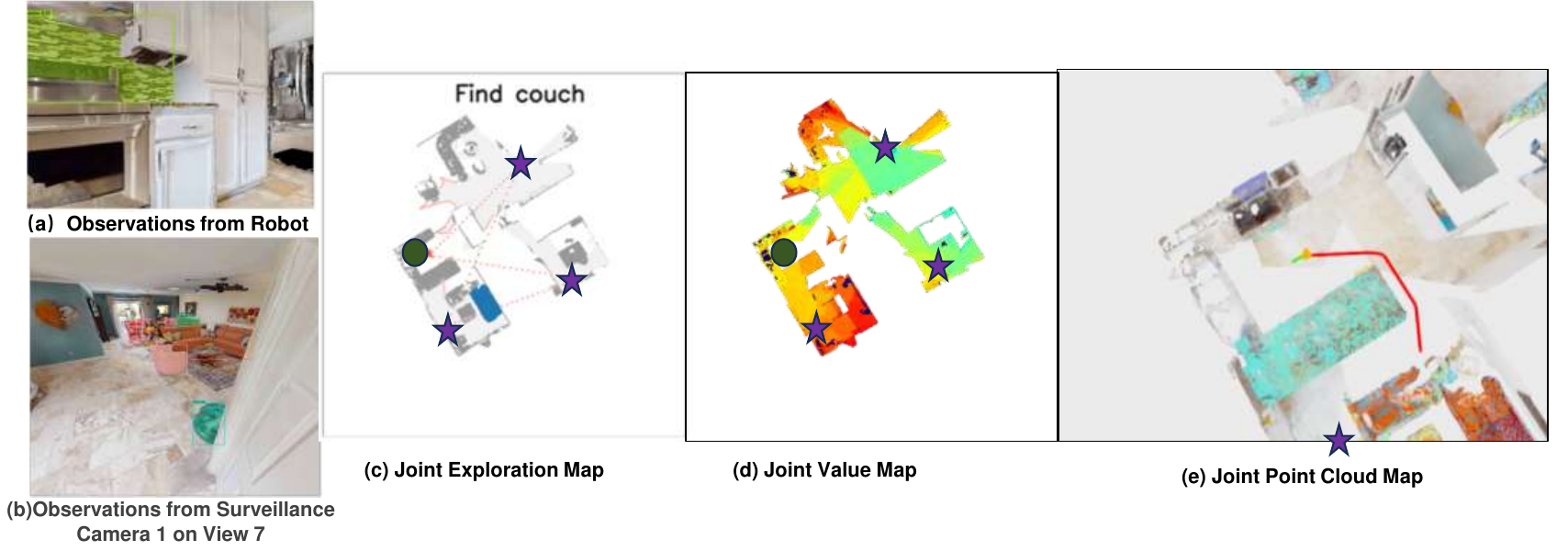}
  \caption{
   The visualization of collaborative navigation in the habitat simulator. Figure (a) and Figure (b) depict the robot's first-person view observations and the observations from the 7th view of surveillance camera 1, respectively. Figure (c) illustrates the joint exploration map of the robot and all surveillance cameras, with the area representing the target region. Figures (d) and (e) display the joint exploration value and the joint point cloud map, respectively. In these figures, red areas denote high-value regions, while in Figure (e), the green lines represent the trajectory already traversed, and the red lines indicate the planned trajectory. The purple pentagram and the green circle represent the positions of the surveillance camera and the robot, respectively.
  }
  \label{fig:sim_vis}
  \vspace{-0.45cm}
\end{figure*}

\textbf{Visualization.}
We visualize the navigation process of \projname \   in Figure~\ref{fig:sim_vis}. 
As shown in Figure~\ref{fig:sim_vis}, the robot (green marker) cannot detect the target due to occlusion by walls. 
However, as illustrated in subfigure (b), Surveillance Camera~1 directly observes the target (couch), allowing the system to set the detected location as the navigation goal. 
This reduces unnecessary exploration and improves navigation efficiency. 
Furthermore, in subfigure (d), the area containing the couch is assigned a high value, demonstrating the effectiveness of our semantic reasoning mechanism.

\subsection{Surveillance Configuration Analysis.}

\begin{table}[h]
\centering
\small
\caption{Impact of the Surveillance Perception Range.}
\label{tab:surveillance_fov}
\begin{tabular}{lccc}
\toprule
\textbf{Surveillance Setup} & \textbf{Perception Range}  & \textbf{SPL} & \textbf{SR}\\
\midrule
% Camera & 180°  & 30.5 & 63.9 \\
Camera & 180°  & 32.1 & 66.8 \\
Panoramic Camera &  360° &  34.5& 67.4  \\
\bottomrule
\end{tabular}
-\vspace{-0.3cm}
\end{table}

\begin{table}[h]
\centering
\small
\caption{Impact of the Surveillance Camera Density}
\label{tab:surveillance_density}
\begin{tabular}{lccc}
\toprule
\textbf{Surveillance Setup} & \textbf{Coverage Density}  & \textbf{SPL}  & \textbf{SR}\\
\midrule
Sparse Coverage & $200m^2$/camera & 33.0  & 66.5\\
Dense Coverage & $100m^2$/camera  & 34.5& 67.4\\
\bottomrule
\end{tabular}
-\vspace{-0.2cm}
\end{table}

\textbf{Impact of the Surveillance Camera Perception Range.} In Table~\ref{tab:surveillance_fov}, we compare the performance of cameras with different perception ranges. We observe that a standard camera with a 180° perception range achieves an SPL of 32.1 and an SR of 66.8, while a panoramic camera with a 360° perception range significantly improves performance, achieving an SPL of 34.5 and an SR of 67.4. This demonstrates that a wider perception range enhances environmental awareness, enabling more accurate navigation and better obstacle avoidance.

\textbf{Impact of the Surveillance Camera Density.}
In Table~\ref{tab:surveillance_density}, we evaluate the effect of camera density on navigation performance. Under sparse coverage ($200m^2$ per camera), the system achieves an SPL of 33.0 and an SR of 66.5, while dense coverage ($100m^2$ per camera) improves performance to an SPL of 34.5 and an SR of 67.4. This indicates that higher camera density provides more comprehensive environmental coverage, leading to more reliable navigation.
Notably, even with dense surveillance, performance bottlenecks persist. This is primarily due to structural occlusions and sub-optimal camera placements that preclude a global viewpoint, as well as erroneous detections from the object perception module which limit further performance improvements.

\subsection{Ablation study of key components.}

\textbf{Impact of Joint Value Map.} To evaluate the effectiveness of different value map construction strategies, we compare greedy search, object-centric, and region-centric approaches, as shown in Table~\ref{tab:value_map}. The greedy search baseline, which lacks semantic understanding, achieves an SR of 0.690 and an SPL of 0.356. The object-centric approach, leveraging CLIP features from detected objects, improves performance slightly with an SR of 0.705 and an SPL of 0.358. However, the region-centric approach, which aggregates semantic information across regions using CLIP, achieves the best results with an SR of 0.711 and an SPL of 0.364. This demonstrates that reasoning about broader regions, rather than individual objects, provides a more comprehensive understanding of the environment, leading to more efficient and successful navigation. These results highlight the importance of incorporating region-level semantic information in value map construction for visual navigation tasks.

\begin{table}[h]
\centering
\small
\caption{Impact of Joint Value Map.}
\label{tab:value_map}
\begin{tabular}{@{}p{2.2cm}p{3.2cm}cc@{}}
\toprule
\textbf{VLM} & \textbf{Strategy}  & \textbf{SPL}& \textbf{SR} \\
\midrule
None & Greedy  & 35.6 & 69.0 \\
CLIP & Object-centric & 35.8 & 70.5 \\
CLIP & Region-centric & 36.4 & 71.1 \\
\bottomrule
\end{tabular}
\vspace{-0.1cm}
\end{table}

\begin{table}[h]
\centering
\small
\caption{Impact of Object Detector.}
\label{tab:component_detector}
\begin{tabular}{>{\raggedright\arraybackslash}p{2.1cm}>{\raggedright\arraybackslash}p{3cm}cc}
\toprule
\textbf{Component} & \textbf{Variant}  & \textbf{SPL}& \textbf{SR} \\
\midrule
Detector & YOLO-World  & 32.65& 59.0 \\
Detector & GroundingDINO & 34.53  & 67.4\\
\bottomrule
\end{tabular}
\vspace{-0.35cm}
\end{table}

\textbf{Impact of Object Detector.} As shown in Table~\ref{tab:component_detector}, we compare two detectors, YOLO-World and GroundingDINO. GroundingDINO achieves superior performance with an SR of 67.4 and an SPL of 34.53, significantly outperforming YOLO-World, which achieves an SR of 59.00 and an SPL of 32.65. This improvement underscores the importance of using advanced detection models like GroundingDINO, which provide more accurate and semantically rich object detections, thereby enhancing the overall navigation system. Together, these results demonstrate that combining region-centric value map construction with a high-quality object detector offers the most effective framework for robust visual navigation tasks.

% 

%%%%%%%%%%%%%%%%%%%%%%%%%%%%%%%%%%%%%%%%%%%%%%%%%%%%%%%%%%%%%%%%%%%%%%%%%%%%%%%%
\section{Conclusions and Limitations}
In this work, we introduced \projname \  , a novel collaborative object navigation framework that leverages both robot egocentric observations and multi-view surveillance inputs. We also constructed the first dataset for robot–surveillance collaboration, spanning 36 scenes, 74 floors, and 206 cameras, providing a systematic benchmark for multi-view navigation. Extensive experiments on HM3D demonstrated that \projname \   achieves state-of-the-art performance in large-scale and multi-floor environments, significantly improving both exploration efficiency and navigation success rate. Ultimately, \projname \   highlights the potential of infrastructure-guided embodied intelligence, opening new directions for real-world applications such as large-scale search, household assistance, and rescue missions.

\textbf{Limitations and Future Work.} A primary constraint is the reliance on precise camera poses for alignment; future research could explore pose-free alignment or visual SLAM to enhance scalability. Furthermore, extending the framework to dynamic environments via active adjustments of surveillance camera FOV and orientation would enable real-time tracking of moving targets. Finally, investigating feature-level compression and adaptive transmission strategies will be crucial to mitigate communication bandwidth and latency during real-world deployment.

\section{ACKNOWLEDGEMENTS}
\vspace{-0.1cm}
\small{
This research is supported by Artificial Intelligence-National Science and Technology Major Project (2023ZD0121200), the National Natural Science Foundation of China (62437001, 62436001, 62441617), the Strategic Priority Research Program of Chinese Academy of Sciences under Grant XDB1350103, the Beijing Natural Science Foundation (L252146), and the Key Research Development Program of Jiangsu Province under Grant BE2023016-3.
}

\vspace{-0.1cm}

\bibliographystyle{IEEEtran}
%\balance
\bibliography{IEEEabrv,References}
\end{document}